\documentclass[10pt]{article}

\usepackage[pdftex]{graphicx}
\graphicspath{{./}}

\usepackage{csagh}
\usepackage{float}
\usepackage{mathtools}
\usepackage{amsfonts}
\usepackage{amsmath}
\usepackage{amsthm}
\usepackage{tikz}
\usepackage{graphicx}
\usepackage{graphics}
\usepackage{subfigure}
\usepackage{epsfig}
\usepackage{notoccite}
\usepackage{booktabs}
\usepackage{multirow}
\usepackage{makecell}

\usepackage[utf8]{inputenc}
\usepackage[T1]{fontenc}

\usepackage{setspace}

\begin{document}
\begin{opening}

\title{Using super-resolution for enhancing visual perception and segmentation performance in veterinary cytology}

\author[ACC Cyfronet AGH, Nawojki 11, 30-950 Krak\'ow, Poland, j.caputa@cyfronet.pl]{Jakub Caputa}

\author[ACC Cyfronet AGH, Nawojki 11, 30-950 Krak\'ow, Poland, AGH University of Science and Technology, al. Mickiewicza 30, 30-059 Krak\'ow, Poland,  wielgosz@agh.edu.pl]{Maciej Wielgosz}
\author[ACC Cyfronet AGH, Nawojki 11, 30-950 Krak\'ow, Poland]{Daria \L{}ukasik}
\author[ACC Cyfronet AGH, Nawojki 11, 30-950 Krak\'ow, Poland, AGH University of Science and Technology, al. Mickiewicza 30, 30-059 Krak\'ow, Poland]{Pawe\l{} Russek}
\author[ACC Cyfronet AGH, Nawojki 11, 30-950 Krak\'ow, Poland]{Jakub Grzeszczyk}
\author[ACC Cyfronet AGH, Nawojki 11, 30-950 Krak\'ow, Poland, AGH University of Science and Technology, al. Mickiewicza 30, 30-059 Krak\'ow, Poland]{Micha\l{} Karwatowski}
\author[ACC Cyfronet AGH, Nawojki 11, 30-950 Krak\'ow, Poland]{Szymon Mazurek}
\author[ACC Cyfronet AGH, Nawojki 11, 30-950 Krak\'ow, Poland, AGH University of Science and Technology, al. Mickiewicza 30, 30-059 Krak\'ow, Poland]{Rafa\l{} Fr\k{a}czek}
\author[University of Life Sciences, al. Akademicka 13, 20-950 Lublin, Poland]{Anna \'S{}miech}
\author[ACC Cyfronet AGH, Nawojki 11, 30-950 Krak\'ow, Poland, AGH University of Science and Technology, al. Mickiewicza 30, 30-059 Krak\'ow, Poland]{Ernest Jamro}
\author[ACC Cyfronet AGH, Nawojki 11, 30-950 Krak\'ow, Poland, AGH University of Science and Technology, al. Mickiewicza 30, 30-059 Krak\'ow, Poland]{Sebastian Koryciak}
\author[ACC Cyfronet AGH, Nawojki 11, 30-950 Krak\'ow, Poland, AGH University of Science and Technology, al. Mickiewicza 30, 30-059 Krak\'ow, Poland]{Agnieszka D\k{a}browska-Boruch}
\author[ACC Cyfronet AGH, Nawojki 11, 30-950 Krak\'ow, Poland, AGH University of Science and Technology, al. Mickiewicza 30, 30-059 Krak\'ow, Poland]{Marcin Pietro\'n}
\author[ACC Cyfronet AGH, Nawojki 11, 30-950 Krak\'ow, Poland, AGH University of Science and Technology, al. Mickiewicza 30, 30-059 Krak\'ow, Poland]{Kazimierz Wiatr}


\begin{abstract}
 
The primary objective of this research was to enhance the quality of semantic segmentation in cytology images by incorporating super-resolution (SR) architectures. An additional contribution was the development of a novel dataset aimed at improving imaging quality in the presence of inaccurate focus. Our experimental results demonstrate that the integration of SR techniques into the segmentation pipeline can lead to a significant improvement of up to 25\% in the mean average precision (mAP) segmentation metric. These findings suggest that leveraging SR architectures holds great promise for advancing the state of the art in cytology image analysis.

\end{abstract}


\end{opening}

\section{Introduction, Motivation}

In recent years, deep learning-based solutions have emerged as a prominent topic in the field of Information Technology. Novel approaches are being developed and implemented daily to optimize, enhance, and facilitate various aspects of our lives. This growing trend is also evident in the medical domain, including veterinary medicine. It is important to note that deploying models for healthcare applications entails significant responsibility and necessitates rigorous testing and monitoring to mitigate any risks associated with artificial intelligence (AI) predictions.

Our research team is dedicated to developing solutions that assist veterinarians in making faster and more accurate diagnoses for their animal patients. Our prior work has focused on age classification \cite{9908709},  object segmentation \cite{grzeszczyk2023segmentation} and detection \cite{app11167181} in the context of cytology imaging for canines. In this study, we continue our investigation into AI applications in the veterinary field, specifically exploring the combination of Super Resolution (SR) and Semantic Segmentation techniques. By building upon previous research, we aim to further advance the state of the art in veterinary image analysis and improve diagnostic outcomes.



\section{State of the art, Reason for conducting the research}

The acquisition of cytology images is a multifaceted process that involves the preparation of tissue samples using staining methods such as Diff-Quik, followed by the selection of suitable areas by a veterinary expert and image capture via a microscope-mounted camera. This study aims to address the challenges associated with obtaining images of inadequate focus or suboptimal quality for examination purposes.

Research Questions:

\begin{enumerate}
\item Can deep learning-based architectures enhance the quality and resolution of cytology images, thereby facilitating improved image quality assessments?

\item To what extent can such enhancements aid pathologists in diagnosing challenging or average-quality cases?

\item Does the improvement of image quality augment the performance of semantic segmentation architectures in detecting carcinogenic cells within preparations?

\item How can a balance be struck between the varying perceptions of machine learning models, metrics, and human evaluators in determining image quality to achieve consistent and reliable results?

\end{enumerate}

\textbf{Dual Super-Resolution Learning for Semantic Segmentation \cite{dual_sr_segm}}

In the presented research authors proposed a two-way framework aimed at enhancing segmentation accuracy without incurring additional computational expenses. Prior investigations have demonstrated that diminishing the resolution of images leads to a decrease in segmentation quality. To address the high training costs associated with larger input sizes, the authors propose integrating super-resolution techniques into semantic segmentation tasks.

The devised architecture can perform segmentation
and super resolution in an end-to-end training paradigm. This  approach has been shown to achieve higher mean Intersection over Union (IoU) metric values while simultaneously requiring less computational power.

\textbf{How Effective Is Super-Resolution to Improve Dense Labelling of
Coarse Resolution Imagery? \cite{how_effective}}

The findings presented in this paper demonstrate that employing super-resolution (SR) techniques effectively enhances semantic segmentation performance. While the proposed approach surpasses conventional interpolation methods, it does not exceed the performance of the original high-resolution data. The developed pipeline features a straightforward design, consisting of two independent modules. The segmentation network receives input from the super-resolved data.

For the selected datasets, it was observed that as the level of degradation increased, the improvement in the Intersection over Union (IoU) metric became more substantial, due to the implementation of the super-resolution module.

\textbf{Simultaneous Super-Resolution and Segmentation Using a Gener-
ative Adversarial Network: Application to Neonatal Brain MRI \cite{mri}}

Super resolution is frequently used in  in the preprocessing of neonatal brain MRI data, owing to the lower resolution obtained in images acquired during examinations. Typically, this process involves a simple upscaling of the image followed by the application of segmentation models.

In this study, the authors introduce a unified solution that leverages a generative adversarial network (GAN). Experimental results demonstrate improved segmentation performance using the proposed method. Furthermore, the authors highlight potential future directions and applications within the field of medical image processing.

\section{Contribution, new algorithm, constructed system}

\begin{figure}[H]
    \centering
    \includegraphics[width=\textwidth]{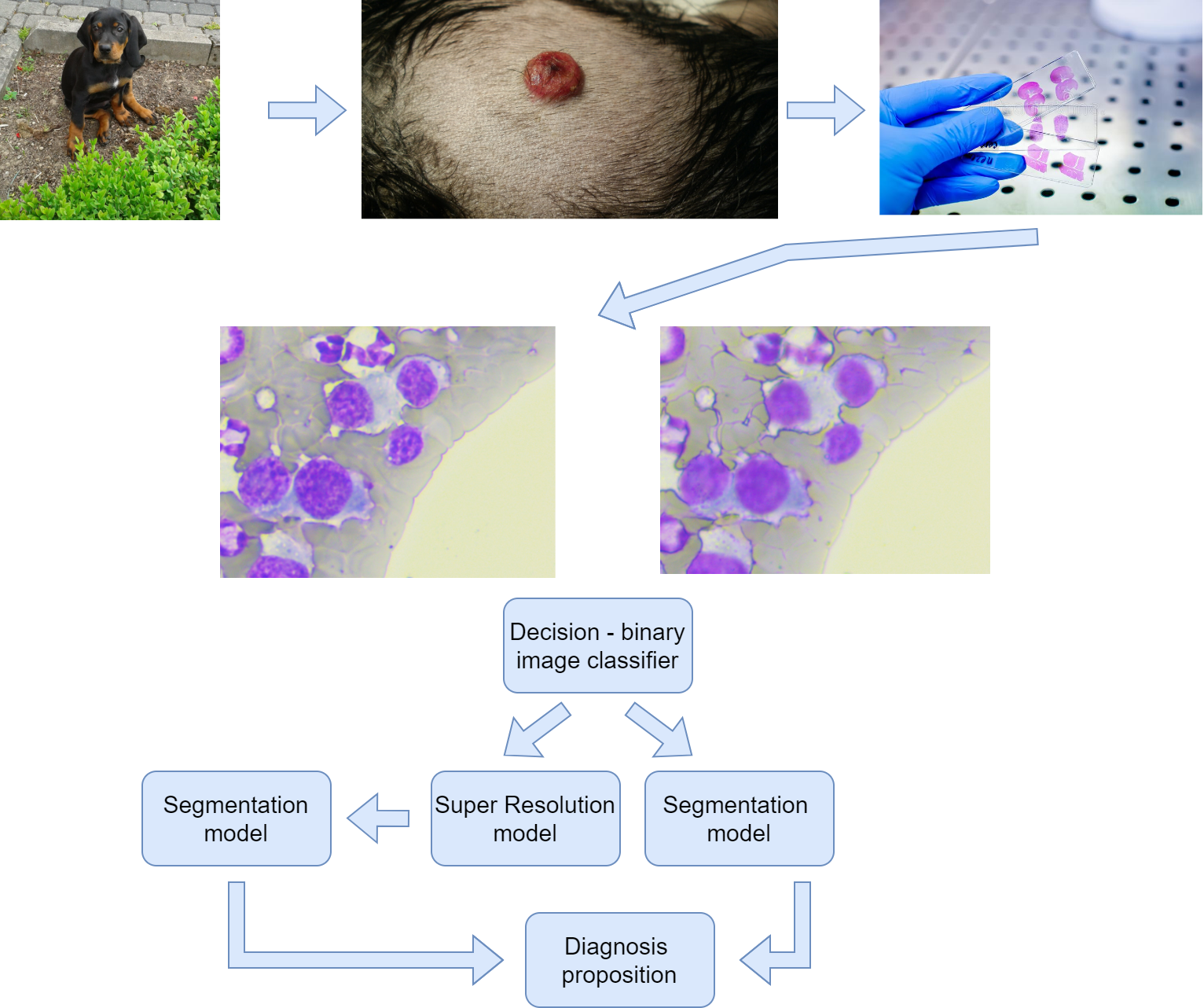}
    \caption{Proposed working system scheme}
    \label{fig:system_scheme}
\end{figure}

The primary contribution of this study is the incorporation of a Super Resolution module into the machine learning pipeline, with the aim of enhancing the accuracy of segmentation models (Figure \ref{fig:system_scheme}). This potential application for improving image quality emerged as a result of various distortions that may occur during the acquisition of cytological preparation images (Table \ref{table:possible_dist}). Our research is specifically focused on addressing poor sharpness distortions.

The objective of this study is to develop a machine learning model capable of enhancing the quality of images affected by improper focus settings on the microscope's adjustment knob. To evaluate the effectiveness of this approach, the enhanced images will be compared to properly created images. The development of a dedicated dataset is a prerequisite for this evaluation \cite{sr_dataset}.

In the context of veterinary examinations, an animal patient undergoes evaluation when visible skin alterations are observed. A tissue sample is subsequently obtained and examined under a microscope, during which an image is generated. In instances where the image quality is sub optimal or the microscope lens focus is improperly set, the decision block (binary image classifier) routes the image to the Super Resolution model. Following this enhancement process, the segmentation model identifies objects within the image, and a diagnosis is proposed.

\subsection{Novel dataset}

The majority of datasets for Super Resolution (SR) tasks are artificially generated, employing image downscaling and interpolation techniques. However, the nature of the distortion we aim to address is distinct from these methods. Recognizing this discrepancy led to the development of an experimental dataset \cite{sr_dataset} in collaboration with a veterinary expert, which is elaborated upon in the subsequent chapter.

The following algorithm was proposed for the acquisition of samples:

\begin{enumerate}
  \item Identify the diagnostic region within the cytological preparation. 
  \item Adjust the microscope lens focus to obtain a high-quality, sharp image.
  \item Intentionally alter the microscope's adjustment knob to degrade the image quality and sharpness, thereby simulating the real-world distortion.
\end{enumerate}

\begin{figure}[H]
\centering
\subfigure[Correct image]{\includegraphics[width=6.3cm]{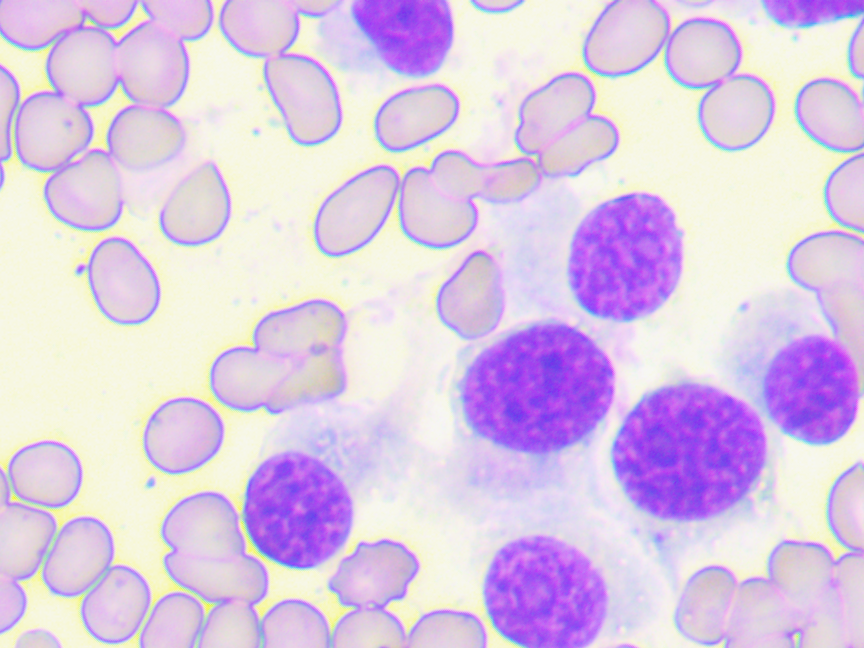}}
\centering
\subfigure[Deliberately distorted image]{\includegraphics[width=6.3cm]{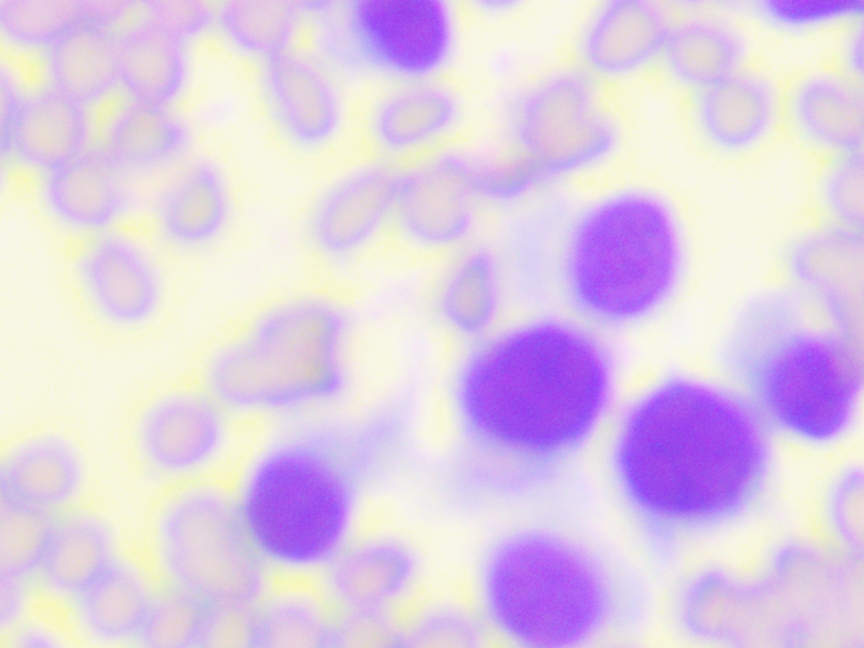}}
\centering
\caption{Examples of high and low sharpness images from the data set}
\label{fig:sample_dataset_image}
\end{figure}

This approach enabled the generation of a dataset  that more accurately represents the specific type of distortion we aim to mitigate, providing a more suitable foundation for model training and evaluation as in Figure \ref{fig:sample_dataset_image}.

Following this procedure we collected 1192 high resolution (2592 $\times$ 1944) samples with their corresponding distorted versions.

\subsection{Proposed new super resolution metric}

\begin{figure}[H]
    \centering
    \includegraphics[scale=0.6]{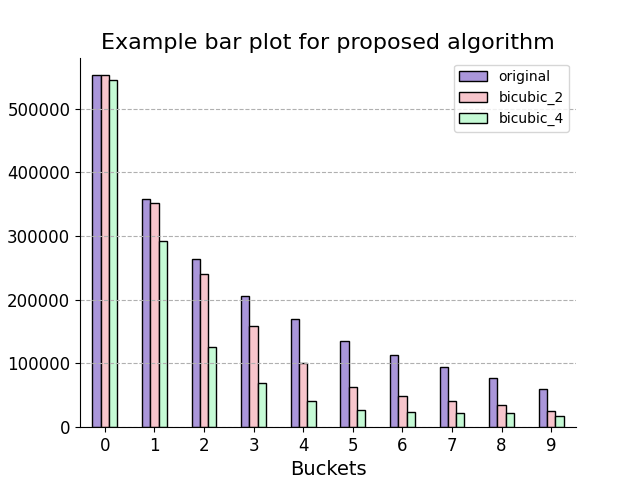}
    \caption{2D-DFT magnitude bar plot for original image and two bicubic interpolation degradations, example with 10 ring-shaped masks}
    \label{fig:dft4}
\end{figure}

In this study, an additional metric based on frequency analysis is employed as a novel approach to assess segmentation performance. This method involves grouping the energy computed using a 2-dimensional Discrete Fourier Transform (DFT-2D) by frequency. This operation enables the observation of the total energy within each frequency range, providing insights into how machine learning models affect high frequencies in the image, which contribute to visual sharpness. The procedure for this approach is as follows (Figure \ref{fig:dft4}):

\begin{enumerate}
  \item Open an image in YCbCr mode and use only luminance channel
  \item Compute the 2-dimensional discrete Fourier Transform
  and shift the zero-frequency component to the center of the spectrum.
  \item Calculate the absolute sum of all magnitudes for a chosen set of ring-shape masks and display the results in a bar plot.
\end{enumerate}


\subsection{Research formula for specific medical use case with unknown degradation}

During the course of the research, addressing the super-resolution (SR) task when the nature of degradation is unknown was a significant concern. The following strategies were assessed:

\begin{enumerate}
  \item Use pre-trained models on various datasets
  \item Investigate known degradations, such as bicubic interpolation on our medical dataset
  \item Develop a dedicated super-resolution dataset exhibiting the same degradation intended to be mitigated.
\end{enumerate}

The most favorable results were achieved using the third approach; however, it is crucial to consider feedback from domain experts in the medical field. It was discovered that applying super-resolution to images introduced artifacts that would not typically be present in cytology images. Veterinary specialists tended to favor lower-quality images over sharper ones, as the artifacts introduced by SR models hindered the diagnostic process. One of the key conclusions drawn is that a sharper image does not necessarily equate to a superior model.


\section{Experiments and results}
This chapter provides a detailed look at the experiments conducted during this research study. Each subchapter explains the different stages of the study in a more accessible way, while still maintaining scientific accuracy.

\subsection{Comparison of possible distortions in cytology imaging}

Table \ref{table:distortion} below presents a list of potential distortions that may occur during image creation by veterinary experts. These hypothetical scenarios may require the application of Super Resolution models as a preprocessing step to recover the images to their desired quality.

\begin{table}[H]
  \centering
  \begin{tabular}{  c  m{5cm}  m{5cm}  }
    \toprule
    \textbf{distortion type} & \textbf{image} & \textbf{description} \\ \midrule
    correct image &
    \begin{center}
    \begin{minipage}{.3\textwidth}
    \centering
      \includegraphics[scale=0.20]{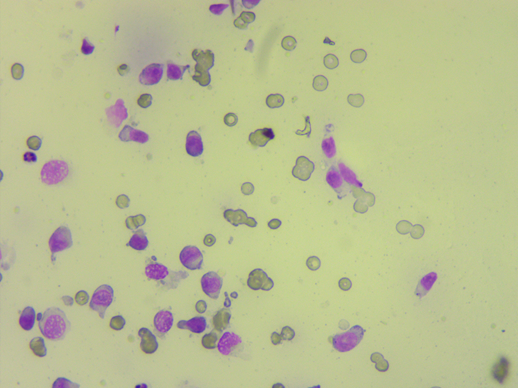}
    \end{minipage}
    \end{center}
    & 
    Image properly created
    \\ \midrule
    dark lighting &
    \begin{center}
    \begin{minipage}{.3\textwidth}
    \centering
      \includegraphics[scale=0.20]{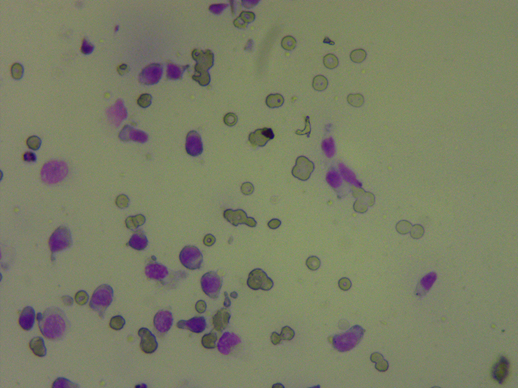}
    \end{minipage}
    \end{center}
    & 
    microscope bulb is not turned on or the room where the image is created is dark, the resulting image may suffer from low contrast and poor illumination.
    \\ \midrule
    
    closed aperture &
    \begin{center}
    \begin{minipage}{.3\textwidth}
    \centering
      \includegraphics[scale=0.20]{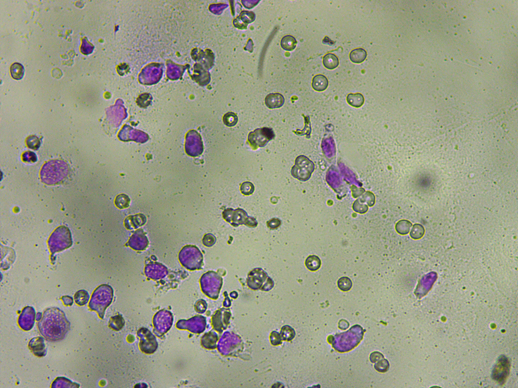}
    \end{minipage}
     \end{center}
    
    & 
    Responsible for the amount of the light that comes to a focus in the image plane
    \\ \midrule
    
    closed condensor &
    \begin{center}
    \begin{minipage}{.3\textwidth}
    \centering
      \includegraphics[scale=0.20]{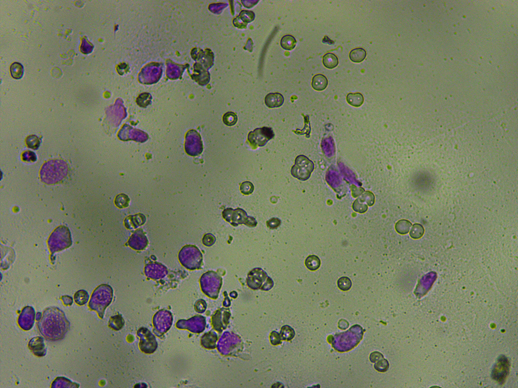}
    \end{minipage}
     \end{center}
    & 
    An improperly adjusted condenser, which is responsible for providing evenly distributed illumination 
    \\ \midrule
    
    dark outside &
    \begin{center}
    \begin{minipage}{.3\textwidth}
    \centering
      \includegraphics[scale=0.13]{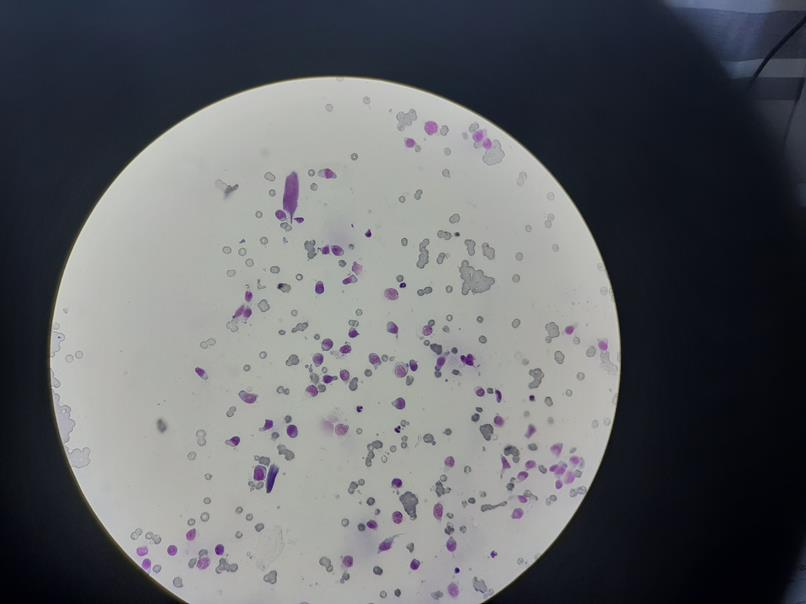}
    \end{minipage}
     \end{center}
    & 
    The image was not directly at the lens leading to dark edges 
    \\ \midrule
    
    bad sharpness &
    \begin{center}
    \begin{minipage}{.3\textwidth}
    \centering
      \includegraphics[scale=0.20]{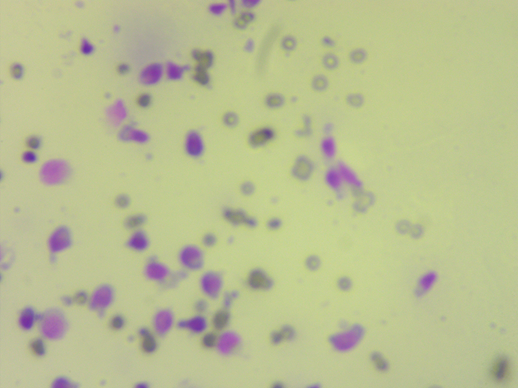}
    \end{minipage}
     \end{center}
    & 
    The microscope screw set inaccurately or the focus is set for the background of the image
    \\ \bottomrule
    
  \end{tabular}
  \caption{Potential Distortions in Veterinary Image Creation}\label{table:possible_dist}
  \label{table:distortion}
\end{table}

\subsection{Comparison of different image upsampling methods using scale factor 2}

\begin{figure}[H]
\centering
\subfigure[Without processing]{\includegraphics[width=4cm]{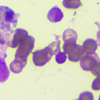}}
\centering
\subfigure[Nearest neighbour ]{\includegraphics[width=4cm]{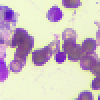}}
\centering
\subfigure[Bilinear interpolation ]{\includegraphics[width=4cm]{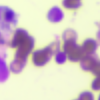}}
\centering
\subfigure[Bicubic interpolation image]{\includegraphics[width=4cm]{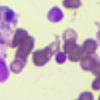}}
\centering
\subfigure[Lanczos 2]{\includegraphics[width=4cm]{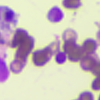}}
\centering
\subfigure[Box shaped kernel]{\includegraphics[width=4cm]{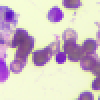}}
\centering
\caption{Comparison of different image upsampling methods using scale 2}
\label{fig:comp_classical}
\end{figure}

This section presents a visual analysis of various classical image upsampling techniques when applied with a scale factor of 2. The objective is to evaluate the performance of each method in terms of image quality, preservation of structural details, and overall effectiveness in enhancing the resolution of the original image (Figure \ref{fig:comp_classical}).



\subsection{Pretrained segmentation model approach}

This section contains the results of inferencing pretrained Super Resolution models and measuring their impact on semantic segmentation task on our cytology dataset.

In this comparative analysis, three state-of-the-art image upsampling architectures were selected for evaluation: SwinIR, BSRGAN and RealSRGAN.The goal was to assess each model's performance in terms of image quality and impact on segmentation metrics.

For segmentation evaluation, a deep learning model based on the Cascade Mask R-CNN \cite{casmrcnn} architecture was selected. The ResNeSt101 \cite{resnest}, which employs skip connections (i.e., input values bypass the current layer without any modifications, and are then summed with the modified input), was used for feature extraction. The model was initialized with weights pre-trained on the MS COCO dataset \cite{coco} and subsequently fine-tuned on cytology images.

\subsubsection{Impact of bicubic interpolation on segmentation inference}

\begin{table}[H]
\centering
\begin{tabular}{lcccccc}
\toprule
\textbf{Factor} & \textbf{segm\_mAP} & \textbf{avg\_precision} & \textbf{avg recall} & \textbf{PSNR} & \textbf{SSIM} & \textbf{LPIPS} \\ 
\midrule
Original                    & 0.439              & 0.623                       & 0.679                   & -             & 1             & 0              \\ 
2                          & 0.392              & 0.598                       & 0.663                   & 34.98         & 0.93          & 0.07           \\ 
3                          & 0.276              & 0.492                       & 0.573                   & 32.62         & 0.82          & 0.17           \\ 
4                          & 0.195              & 0.444                       & 0.540                   & 32.19         & 0.79          & 0.26           \\ 
5                          & 0.113              & 0.348                       & 0.456                   & 31.68         & 0.74          & 0.35           \\ 
\bottomrule
\end{tabular}
\caption{Comparison of super resolution and segmentation metrics using pretrained segmentation model for inference}
\label{table:results_bicubic}
\end{table}

\begin{figure}[H]
    \centering
    \includegraphics[scale=0.5]{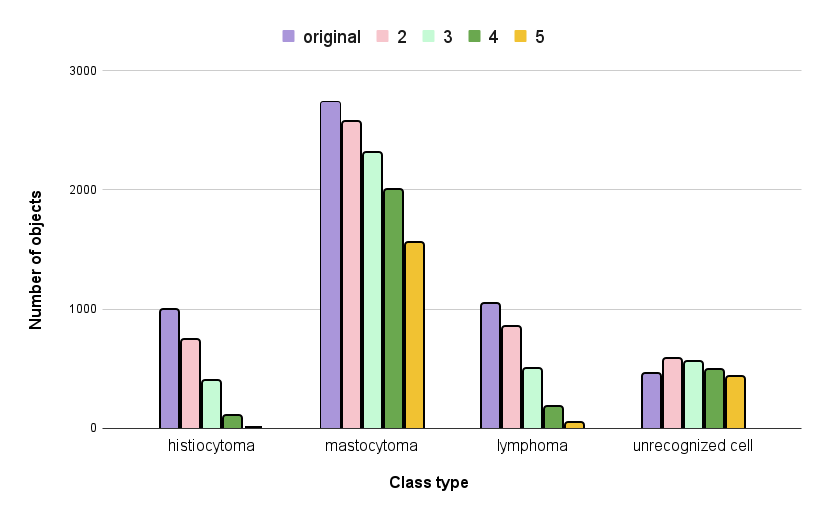}
    \caption{Number of True Positive cells detection for different scaling factors}
    \label{fig:tp}
\end{figure}

The results presented in Table \ref{table:results_bicubic} reveal a expected trend. As the bicubic interpolation scaling factor increases, which corresponds to a greater loss of information, both segmentation and super-resolution metrics are negatively affected. A higher scaling factor leads to increased confusion between objects. For instance, with bicubic interpolation using a scaling factor of 5, almost no cells are accurately recognized for the two cancer types, as illustrated in Figure \ref{fig:tp}.

\begin{figure}[H]
    \centering
    \includegraphics[scale=0.6]{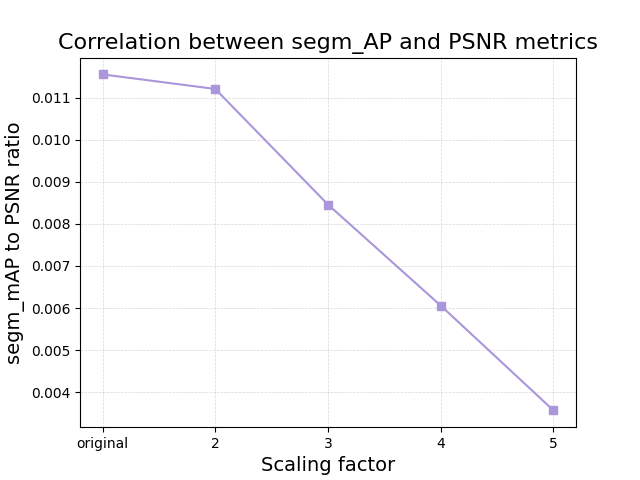}
    \caption{Correlation between segmAP and PSNR metrics for different scaling factors}
    \label{fig:segmpsnr}
\end{figure}

Figure \ref{fig:segmpsnr} presents the relationship between the segmentation Average Precision (segmAP) and Peak Signal-to-Noise Ratio (PSNR) metrics. A decreasing trend in the ratio between these two metrics is observed when the scaling factor increases. This indicates that, in some cases, a linear correlation exists between the performance metrics of these two distinct computer vision tasks. As the scaling factor increases, the quality of the image and segmentation performance both tend to degrade.


\subsubsection{Impact of the pre-trained Super Resolution models on segmentation inference}

In this phase, a comparative analysis is conducted to evaluate the performance of the segmentation model on the original dataset opposite to bicubic interpolation with scaling factors of 2 and 5. The primary objective is to investigate whether employing pre-trained models can enhance the accuracy of the segmentation process.


\begin{table}[H]
\centering
\begin{tabular}{lcccccc}
\toprule
\textbf{Data Set Original} & \textbf{segm\_mAP} & \textbf{avg\_precision} & \textbf{avg\_recall} & \textbf{PSNR} & \textbf{SSIM} & \textbf{LPIPS} \\ 
\midrule
\textbf{Original}          & 0.457     & 0.596              & 0.654           & -    & -    & -     \\ 
RealSRGAN                  & 0.446              & 0.587                       & 0.540                    & 36.16         & 0.96          & 0.050          \\ 
BSRGAN                     & 0.430              & 0.572                       & 0.636                    & 33.54         & 0.93          & 0.092          \\ 
SwinIR                     & 0.421              & 0.567                       & 0.630                    & 35.10         & 0.96          & 0.057          \\ 
SwinIR\_large              & 0.457              & 0.584                       & 0.643                    & 37.18         & 0.977         & 0.041          \\ 
\bottomrule
\end{tabular}
\caption{Results for inference on the test set for the original data set}
\label{table:inference_seg}
\end{table}

The application of various super-resolution (SR) architectures on the original data did not yield any improvements in segmentation quality. Nevertheless, the minimal loss in mean average precision (mAP) indicates that the model is proficient in identifying cancer cells that have undergone enhancement through the SR process (Table \ref{table:inference_seg}).

When employing image enhancement techniques for damaged data with decimation and bicubic interpolation, the results, as presented in Table \ref{table:bic2} and Table \ref{table:bic5}, are found to be worse. This suggests that the utilization of SR introduces additional noise to the data, consequently leading to poorer performance by the segmentation model. This outcome was anticipated, given that the model was pre-trained on original data and was subsequently required to handle processed data during the inference phase.

\begin{figure}[H]
\centering
\subfigure[Distortion]{\includegraphics[width=4cm]{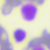}}
\centering
\subfigure[Good quality]{\includegraphics[width=4cm]{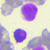}}
\centering
\subfigure[BSRGAN]{\includegraphics[width=4cm]{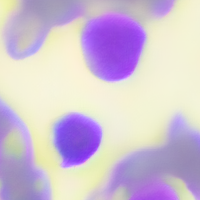}}
\centering
\subfigure[realESRGAN]{\includegraphics[width=4cm]{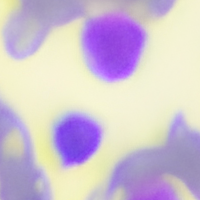}}
\centering
\subfigure[SwinIR]{\includegraphics[width=4cm]{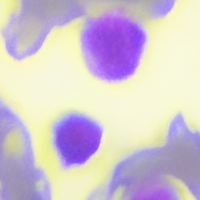}}
\centering
\subfigure[SwinIR large]{\includegraphics[width=4cm]{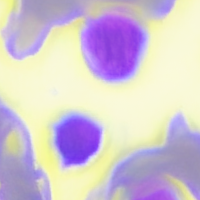}}
\centering
\caption{Example results of chosen architectures (cropped part of the image)}
\label{fig:pred}
\end{figure}

The findings indicate that the naive application of SR models to images does not yield improvements in segmentation quality. However, it does enhance the perceptual quality of the image, as depicted in Figure \ref{fig:pred}. In certain instances, it also results in an increase in SR metrics, as demonstrated in Table \ref{table:inference_seg}.

The obtained results were subsequently reviewed in consultation with a veterinary expert. It was determined that, although the images appeared sharper following the application of the super-resolution model, the presence of certain artifacts rendered them less reliable than the original images. The enhanced image quality did not contribute to improved diagnostic accuracy, as these artifacts introduced elements that would not typically be found in cytology images.

\subsubsection{Training semantic segmentation model on Super-Resolution medical data}
\renewcommand{\arraystretch}{1.5}
\begin{table}[H]
\centering
\begin{tabular}{l>{\raggedright\arraybackslash}m{5cm}cccc}
\toprule
\textbf{Data set} & \textbf{Data set explanation}                                                                                     & \textbf{segm\_mAP} & \textbf{avg\_precision} &

\textbf{avg\_recall} \\ 
\midrule
original          & Original cancer inflammation dataset                                                                             & \textbf{0.494}     & \textbf{0.494}          & 0.584                \\ 
bicubic2          & Original dataset decimated and interpolated using scaling factor 2 (half of the pixels left after decimation)    & 0.465              & 0.465                   & 0.567                \\ 
bicubic4          & Original dataset decimated and interpolated using scaling factor 4 (quarter of the pixels left after decimation) & 0.423              & 0.423                   & 0.527                \\ 
realSRGAN x4      & Original dataset upscaled using realSRGAN and resized to original size                                            & 0.487              & 0.487                   & \textbf{0.587}       \\ 
BSRGAN x4         & Original dataset upscaled using BSRGAM and resized to original size                                               & 0.478              & 0.478                   & 0.587                \\ 
SwinIR x4         & Original dataset upscaled using SwinIR and resized to original size                                              & 0.482              & 0.482                   & 0.579                \\ 
\bottomrule
\end{tabular}
\caption{Training semantic segmentation on interpolated and super resolution data}
\label{table:train_seg}
\end{table}
\renewcommand{\arraystretch}{1}

In this experiment, we trained and tested the segmentation on data processed in various ways. As demonstrated in Table \ref{table:train_seg}, we downsampled the images using decimation and bicubic interpolation and then upsampled using super resolution architectures. While this research may not hold practical significance from a medical standpoint, as manipulating original data is generally discouraged, it does reveal that the optimal results for cancer cell recognition are obtained when utilizing undistorted, original data.

\subsubsection{Dedicated data set experiments for super resolution}

Ultimately, the experiments were carried out on a dedicated dataset, with the SwinIR architecture selected for training \cite{swin_transformer}.

\begin{figure}[H]
\centering
\subfigure[Bicubic interpolation histogram]{\includegraphics[width=6cm]{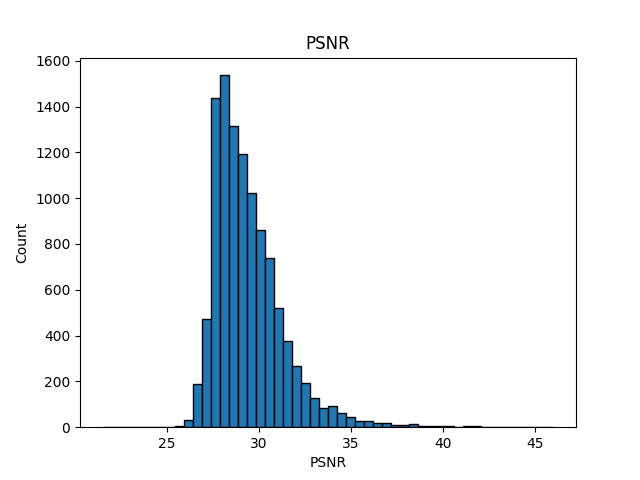}}
\centering
\subfigure[Dedicated data set histogram]{\includegraphics[width=6cm]{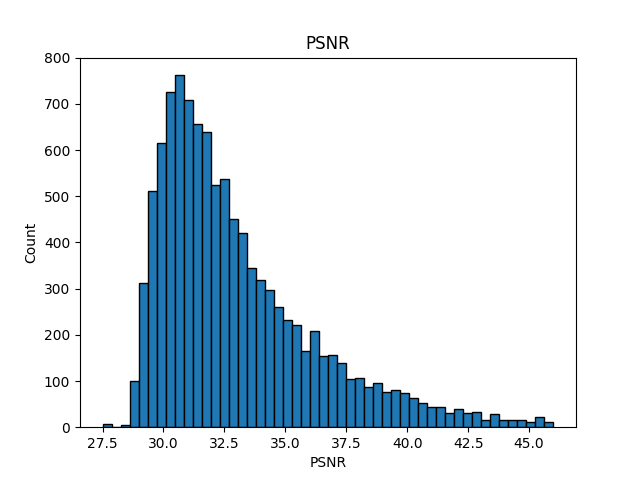}}
\centering

\caption{Comparison of histograms for data sets}
\label{fig:hist_dedicated_dataset}
\end{figure}

During the exploration of the dataset, the data distribution was analyzed. The histograms presented in Figure \ref{fig:hist_dedicated_dataset} showcase the PSNR values for both bicubic interpolation and our dedicated dataset, in comparison to high-resolution original data. The spectrum of our data set is wider and there are images that would be considered of a good quality in terms of pixel loss. In contrast, bicubic interpolation exhibits less diversity, limiting its applicability to the restoration of specific distortion types.

The dedicated dataset encompasses various forms of degradation that are likely to be encountered in cytology images.



\begin{figure}[H]
\centering
\subfigure[Good quality]{\includegraphics[width=4.2cm]{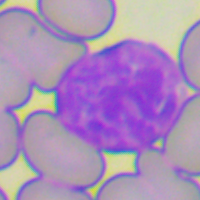}}
\centering
\subfigure[Distortion]{\includegraphics[width=4.2cm]{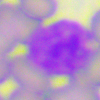}}
\centering
\subfigure[Dedicated start]{\includegraphics[width=4.2cm]{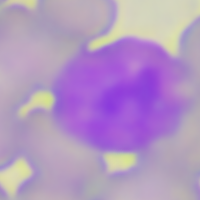}}
\centering
\subfigure[Dedicated stop]{\includegraphics[width=4.2cm]{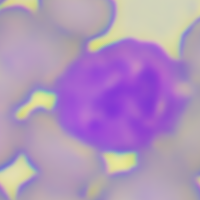}}
\centering
\subfigure[Bicubic start]{\includegraphics[width=4.2cm]{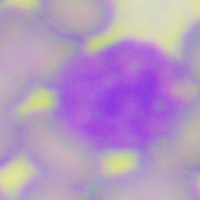}}
\centering
\subfigure[Bicubic stop]{\includegraphics[width=4.2cm]{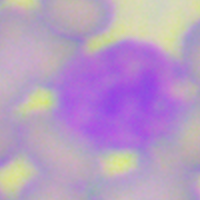}}
\centering
\caption{Comparison of results for training on dedicated and bicubic data sets}
\label{fig:results_dedicated_train}
\end{figure}

The initial experiment exhibited a substantial improvement in the PSNR metric upon training the SwinIR model on our dataset, as displayed in Table \ref{table:tab_res}. This improvement is also evident in the inferred images after training, presented in Figure \ref{fig:results_dedicated_train}.

The transformer model underwent training for approximately 1,000 epochs, utilizing four NVIDIA V-100 GPUs from the ACK Cyfronet Prometheus supercomputer \cite{cyfronet}. Default parameters tailored for the classical super-resolution task were employed. The upsampling factor was set to 2.



Experiments involving training on our data set demonstrated a notable improvement when compared to the bicubic data set, which is commonly utilized in super-resolution tasks.

\begin{figure}[H]
    \centering
    \includegraphics[scale=0.3]{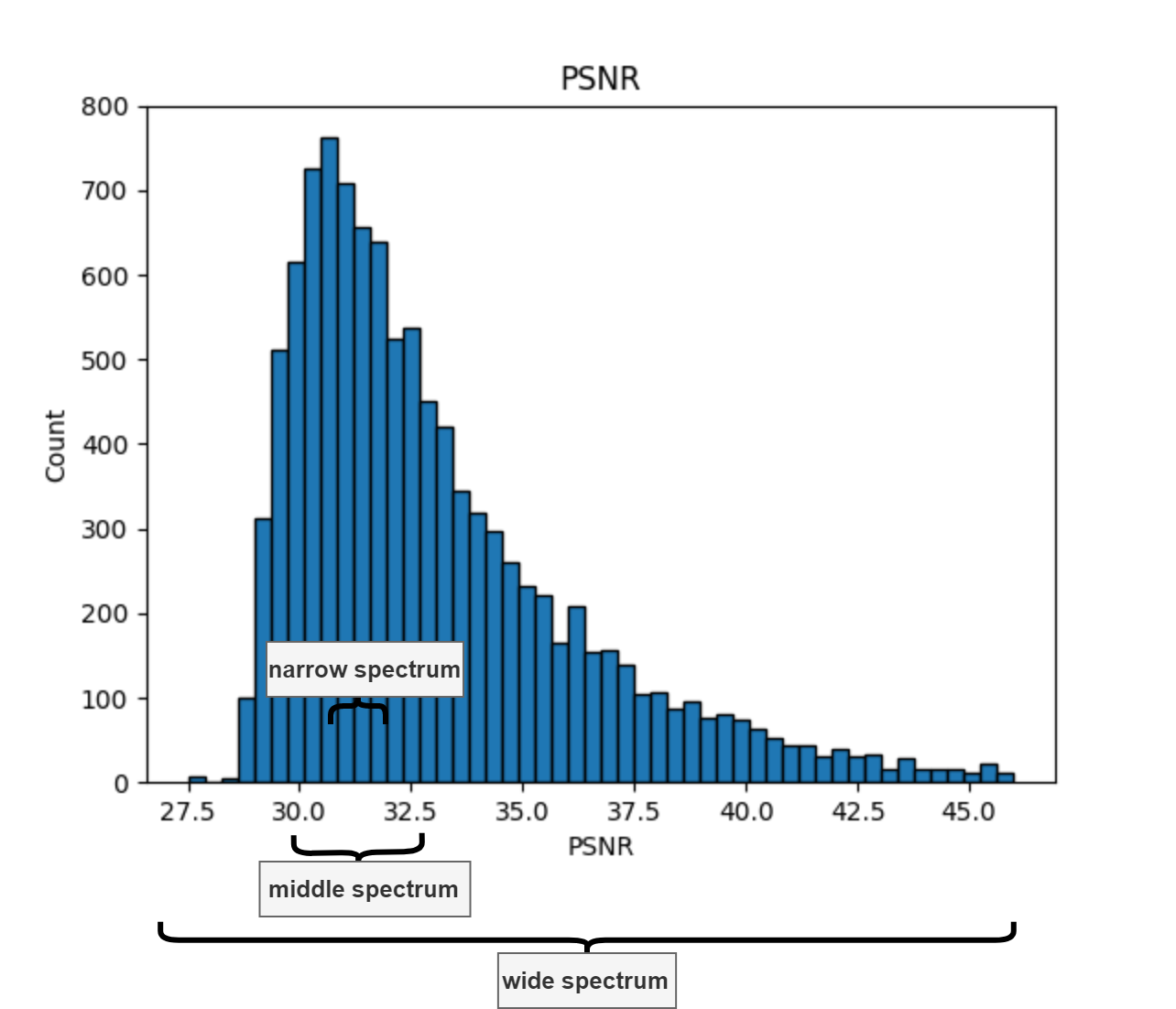}
    \caption{Histogram presenting chosen spectrums}
    \label{fig:spect_hist_expl}
\end{figure}

The second experiment focused on examining the influence of the selected spectrum. We investigated whether a narrow or wide spectrum of our data set would yield superior results. Figure \ref{fig:spect_hist_expl} illustrates the three distinct segments of the dataset that were employed in this experiment.

\begin{figure}[H]
\centering
\subfigure[Good quality]{\includegraphics[width=4.2cm]{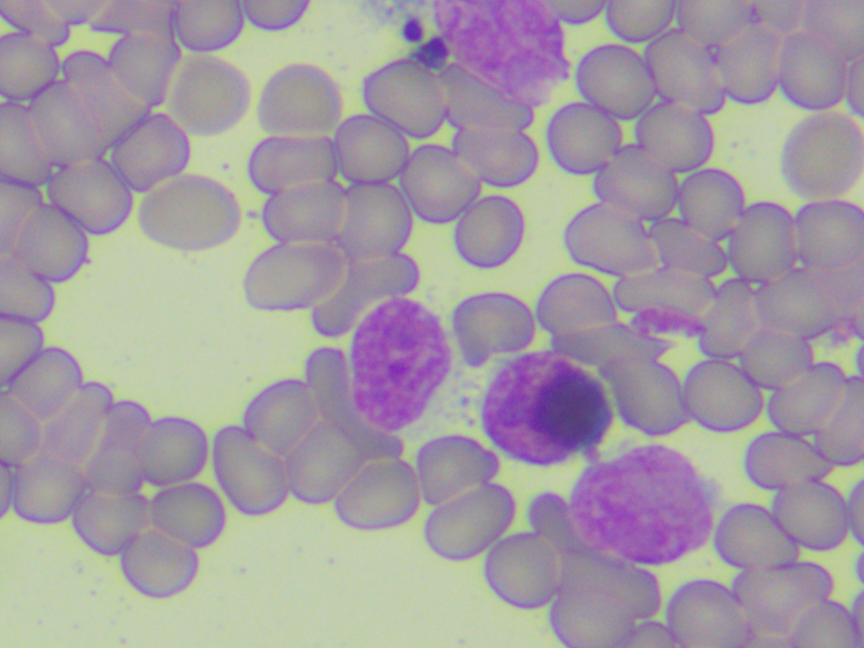}}
\centering
\subfigure[Distortion]{\includegraphics[width=4.2cm]{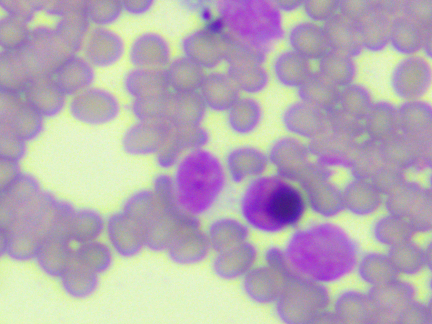}}
\centering
\subfigure[Result]{\includegraphics[width=4.2cm]{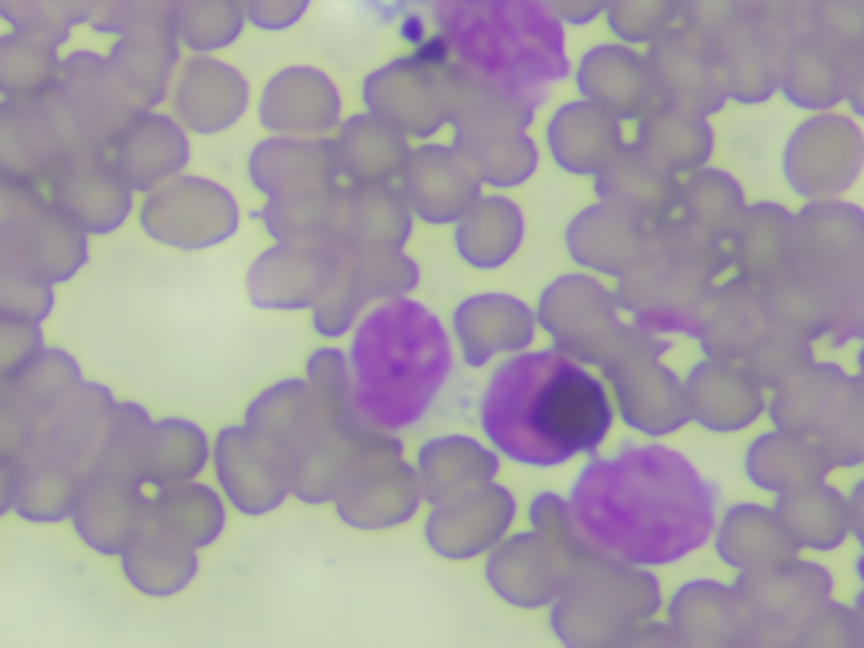}}
\centering
\caption{Example result for training on a wide spectrum}
\label{fig:inf}
\end{figure}

The experiment reveals that training on the widest spectrum (as illustrated in Figure \ref{fig:spect_hist_expl}) leads to the the most favorable results for both narrow and wide spectrum test data sets, as presented in Table \ref{table:spectrum_results}. This finding suggests that the model effectively learns to reconstruct images when the training data encompasses a diverse range of distortions with varying types and levels (Figure \ref{fig:inf}).

\subsubsection{Improving segmentation metrics results with super resolution}

For end-to-end experiments aimed at enhancing segmentation using super-resolution (SR) architectures, a new subset of the dataset was employed. This subset comprised both semantic segmentation annotations and distorted images representing the exact same area.

\begin{table}[H]
\centering

\begin{tabular}{cccccc}

\toprule
\textbf{\makecell{Experiment\\ summary}} & \textbf{\makecell{segm\_\\mAP\_50}} & \textbf{\makecell{segm\_\\mAP\_75}} & \textbf{\makecell{segm\_mAP\_50\\ percent\_change}} & \textbf{\makecell{segm\_mAP\_75\\ percent\_change}} \\ 
\midrule
high quality data set & \textbf{0.314}        & \textbf{0.255} &38.94\% &	63.46\%        \\ 
low quality data set  & 0.226        & 0.156 &0.00\% &	0.00\%        \\ 
subsampling x2, SR x2  & 0.198        & 0.120 & -12.39\% &	-23.08\%        \\ 
SR x2, subsampling x2  & 0.233        & 0.110 & 3.10\% &	-29.49\%        \\ 
subsampling x4, SR x4  & 0.227        & 0.155 &0.44\% &	-0.64\%       \\ 
SR x4, subsampling x4 & \textbf{0.279}        & \textbf{0.195} & 23.45\% &	25.00\%        \\ 
\bottomrule
\end{tabular}
\caption{End-to-end pipeline tests BSRGAN}
\end{table}

The results depicted in the subsequent table exhibit promising outcomes. In certain methodologies, the results show improvements when compared to the low-quality dataset. For instance, with the BSRGAN architecture, employing a 4x upsampling technique followed by subsampling to the required resolution, the improvement reaches up to 25\% when compared to the results that would have been obtained using a low-quality dataset. As anticipated, the most optimal results are achieved when training the segmentation on high-quality images, where the segm\_mAP\_75 increase is up to 64\%. Interesting aspect of the following experiments is that using super resolution as a first step, before subsampling to desired annotations size leads to better results than the opposite operation. This is because subsampling as a first step loses even more informations that can not be restored during upsampling step, leading to lower segmentation results.

The intriguing outcome of the subsequent experiments suggests that implementing super-resolution as an initial step, prior to the subsampling process to achieve the desired annotation size, yields superior results compared to the inverse sequence of operations. This observed phenomena could be primarily attributed to the fact that the adoption of subsampling as a preliminary step entails an excessive loss of information. This loss of detail, once occurred, is unable to be fully recuperated during the upsampling process, culminating in compromised segmentation outcomes. Hence, the order of these processing steps plays a critical role in maximizing the data fidelity and overall accuracy of the segmentation results.

\section{Conclusion}

The presented research, which encompasses two fields of Computer Vision - Super Resolution and Semantic Segmentation, underscores the possibility of enhancing the quality of medical images for their interpretation and analysis. The primary challenges identified during the investigation include a scarcity of data and differing perceptions between programmers and medical experts.

The first challenge stems from the nature of the formulated problem. Restoring an image from an unknown degradation is a daunting task, particularly in medical imaging. Consequently, a unique dataset was created for this study, enabling improvements in both super-resolution metrics and human perception. The second challenge arises from the specific use case provided; veterinary experts analyze medical images differently from individuals unfamiliar with cytology. The realization that sharper, high-resolution images are not always preferable for diagnosis due to artifacts observable after applying AI models was not immediately evident.

This research primary achievements include formulating a potential procedure for addressing unknown degradation, such as incorrectly set sharpness on a microscope. The steps taken during the experimental phase could potentially be applied to other domain-specific use cases. Another valuable aspect is the identification and analysis of possible distortions in medical imaging. We facilitated a better understanding of the problem's nature and its uniqueness compared to standard Super Resolution tasks.

Finally as expected, the substantial increase in the PSNR measure during SwinIR architecture training (Table \ref{table:tab_res}) and the visual perception improvement shown in Figure \ref{fig:results_dedicated_train} are noteworthy. The remarkable improvement of up to 25\% in certain experimental scenarios, with respect to the segm\_mAP metric change, is also worth mentioning. This underscores the potential of the applied methods in enhancing the performance of image segmentation tasks in medical imaging.


\section*{Appendix A. Supplementary material}

\begin{table}[H]
\centering
\begin{tabular}{lcccccc}
\toprule
\textbf{Data set bicubic2} & \textbf{segm\_mAP} & \textbf{avg\_precision} & \textbf{avg recall} & \textbf{PSNR}  & \textbf{SSIM} & \textbf{LPIPS} \\ 
\midrule
\textbf{original}          & \textbf{0.426}     & \textbf{0.566}              & \textbf{0.633}           & \textbf{34.69} & \textbf{0.92} & 0.0084         \\ 
realSRGAN                  & 0.379              & 0.559                       & 0.624                    & 34.49          & 0.92          & \textbf{0.073} \\ 
BSRGAN                     & 0.360              & 0.536                       & 0.607                    & 33.77          & 0.90          & 0.090          \\ 
SwinIR                     & 0.345              & 0.534                       & 0.606                    & 33.54          & 0.91          & 0.087          \\ 
SwinIR\_large              & 0.339              & 0.541                       & 0.613                    & 33.70          & 0.91          & 0.091          \\ 
\bottomrule
\end{tabular}
\caption{Results for inference on the test set for bicubic interpolation with factor 2}
\label{table:bic2}
\end{table}

\begin{table}[H]
\centering
\begin{tabular}{lcccccc}
\toprule
\textbf{Data set bicubic5} & \textbf{segm\_mAP} & \textbf{avg\_precision} & \textbf{avg recall} & \textbf{PSNR}  & \textbf{SSIM} & \textbf{LPIPS} \\ 
\midrule
\textbf{original}          & \textbf{0.116}     & \textbf{0.311}              & \textbf{0.419}           & \textbf{31.58} & \textbf{0.72} & \textbf{0.377} \\ 
realSRGAN                  & 0.092              & 0.300                       & 0.397                    & 31.38          & 0.69          & 0.252          \\ 
BSRGAN                     & 0.093              & 0.276                       & 0.381                    & 31.46          & 0.70          & 0.282          \\ 
SwinIR                     & 0.071              & 0.250                       & 0.374                    & 31.29          & 0.69          & 0.301          \\ 
SwinIR\_large              & 0.055              & 0.227                       & 0.357                    & 31.17          & 0.67          & 0.348          \\ 
\bottomrule
\end{tabular}
\caption{Results for inference on the test set for bicubic interpolation with factor 5}
\label{table:bic5}
\end{table}

\begin{table}[H]
\centering
\begin{tabular}{ccc}
\toprule
\textbf{Train Data Set} & \textbf{Test Data Set} & \textbf{PSNR} \\ 
\midrule
bicubic                 & dedicated              & 25.84         \\ 
dedicated               & dedicated              & \textbf{29.99} \\ 
\bottomrule
\end{tabular}
\caption{Dedicated and bicubic results}
\label{table:tab_res}
\end{table}

\begin{table}[h]
\centering
\renewcommand{\arraystretch}{1.5}
\begin{tabular}{ccccc}
\toprule
\multicolumn{5}{c}{\textbf{Trained}} \\ 
\midrule
\multirow{3}{*}{\textbf{Validated}} & \textbf{Spectrum} & \textbf{Narrow} & \textbf{Middle} & \textbf{Wide}  \\ 
\cmidrule{2-5}
& \textbf{Narrow}   & 30.41           & \textbf{30.14}  & 29.73          \\ 
& \textbf{Middle}   & 30.64           & 30.06           & 29.69          \\ 
& \textbf{Wide}     & \textbf{30.76}  & 29.40           & \textbf{29.90} \\ 
\bottomrule
\end{tabular}
\caption{Spectrum experiments PSNR value}
\label{table:spectrum_results}
\end{table}


\section*{Declaration of interest}
The authors declare that they have no known competing financial interests or personal relationships that could have appeared to influence the work reported in this paper.

\bibliographystyle{cs-agh}
\bibliography{bibliography}

\end{document}